\documentclass[10pt,twocolumn,letterpaper]{article}

\usepackage{iccv}              

\usepackage{float} 
\usepackage{xcolor}
\definecolor{baseline}{RGB}{255,255,180}  

\definecolor{iccvblue}{rgb}{0.21,0.49,0.74}
\usepackage[pagebackref,breaklinks,colorlinks,allcolors=iccvblue]{hyperref}

\def\paperID{20} 
\def\confName{ICCV}
\def\confYear{2025}

\title{From General to Specialized: The Need for Foundational Models in Agriculture}

\author{
Vishal Nedungadi$^{1}$\thanks{Equal contribution. The order of equal-contributing authors was determined randomly using \texttt{random.shuffle} with seed 107.} \quad
Xingguo Xiong$^{1,2*}$ \quad 
Aike Potze$^{1*}$ \quad
Ron Van Bree$^1$\\
Tao Lin$^2$ \quad
Marc Rußwurm$^1$ \quad 
Ioannis N. Athanasiadis$^1$ \\
{\tt\small $^1$ Wageningen University and Research} \\
{\tt\small $^2$ Zhejiang University}
}

\begin{document}
\maketitle
\begin{abstract}
Food security remains a global concern as population grows and climate change intensifies, demanding innovative solutions for sustainable agricultural productivity. 
Recent advances in foundation models have demonstrated remarkable performance in remote sensing and climate sciences, and therefore offer new opportunities for agricultural monitoring. However, their application in challenges related to agriculture—such as crop type mapping, crop phenology estimation, and crop yield estimation—remains underexplored. In this work, we quantitatively evaluate existing foundational models to assess their effectivity for a representative set of agricultural tasks. From an agricultural domain perspective, we describe a requirements framework for an ideal agricultural foundation model (CropFM). We then survey and compare existing general-purpose foundational models in this framework and empirically evaluate two exemplary of them in three representative agriculture-specific tasks. Finally, we highlight the need for a dedicated foundational model tailored specifically to agriculture.
\end{abstract}

\section{Introduction}
\label{sec:intro}

Sustainable agriculture is essential to address the global challenge of eradicating hunger, as emphasized in Sustainable Development Goal 2. With the global population continuing to grow, meeting the rising food demand becomes increasingly important \cite{vanDijk2021}. Traditional agricultural intensification through land use changes can boost crop production but lead to significant environmental risks. A more sustainable solution lies in increasing crop productivity \cite{Rezaei2023}. This calls for advanced strategies that enhance crop productivity through data-driven crop management and decision-making. 

Earth observation (EO) and multi-source remote sensing have become promising tools for enabling sustainable agricultural practices. EO data provide large-scale and timely insights into crop growth and land use conditions, forming the foundation for data-driven management of crop production systems \cite{Benami2021, Burke2021}. Multi-source spatial, temporal, and spectral imagery from remote sensing enables the monitoring of essential crop information such as crop type and crop phenology. Crop type maps detect the spatial allocation patterns of cropping systems, while crop phenology captures the dynamic progression of crop growth, both of which are important to accurately monitoring crop yield and productivity \cite{lin2021earlyinseasoncroptype, Ru_wurm_2020,Xu2020, Graf2023, Shen2023, Lobell2015}. Furthermore, integrating remote sensing data with climate variables such as temperature and precipitation enhances our ability to assess crop-environment interactions. 

Process-based crop models use environmental and agronomic inputs for biophysical process simulation but struggle to integrate multi-modal remote sensing data across large scales \cite{Shuai2022}. Here, the rapid advancement of data-driven learning and AI technology has enabled processing these remote sensing and earth observation data for crop monitoring tasks at large scale. Deep learning models can learn spatial-temporal-spectral patterns concealed within raw remote sensing data to capture dynamism of the agricultural system with enhanced precision\cite{jiang_deep_2020, You2017}. Recent advances have led to the surge of foundation models (FMs), which are pre-trained on large-scale heterogeneous data using self-supervised methods and are capable of generalizing across diverse downstream tasks \cite{chang2025generalizabilityfoundationmodelscrop}. Despite their growing success in remote sensing and climate domains \cite{Cong2022, tseng2023lightweight, nedungadi2024mmearthexploringmultimodalpretext, Bi2023, bodnar2024foundationmodelearth}, their direct application to sustainable agriculture remains underexplored. This paper investigates the application of foundation models to three representative agricultural tasks: crop type mapping, crop phenology estimation and crop yield estimation. Specifically, our contributions
can be summarized as follows:\begin{enumerate}
    \item We introduce a framework or requirements to define the characteristics of an ideal foundation model for agriculture, nicknamed CropFM.
    \item We map existing foundation models within this proposed framework.
    \item We empirically evaluate two such foundation models on the tasks of crop type mapping, crop phenology estimation, and crop yield estimation.
\end{enumerate}

\begin{table*}[!tbh]
\centering
\renewcommand{\arraystretch}{1.0}
\begin{tabular}{p{2.2cm}|p{1.7cm}p{1.5cm}|p{1.5cm}p{1.5cm}|p{0.3cm}p{0.3cm}p{0.6cm}|p{0.3cm}p{0.3cm}p{0.3cm}|p{0.3cm}p{0.3cm}p{0.3cm}}
\toprule

\textbf{Model} & 
\multicolumn{2}{c|}{\textbf{Temporal}} &
\multicolumn{2}{|c|}{\textbf{Spatial}} &
\multicolumn{3}{|c|}{\textbf{RS}} &
\multicolumn{3}{|c|}{\textbf{Meteo}} &
\multicolumn{3}{|c}{\textbf{Other}} \\
& Extent & Resolution & Extent & Resolution & MS & VI & SAR & T & P & + & S & M & E \\
\midrule
\midrule
Ideal CropFM & $\geq$ 1 year & $\leq$ 1 day & $\geq$ 1 km$^2$ & $\leq$ 10m& \checkmark  & \checkmark & \checkmark & \checkmark & \checkmark & \checkmark& \checkmark & \checkmark & \checkmark \\
\midrule
MMEarth & - & - & 0.4km$^2$ & 10m   & \checkmark  & \checkmark & \checkmark & \checkmark & \checkmark && & & \\
AnySat &140 weeks&1 week&1.6km$^2$  &0.2m&\checkmark&& \checkmark&&&&&&\\ 
Galileo &2 years&1 month&0.9km$^2$&10m  &\checkmark&\checkmark & \checkmark&\checkmark&\checkmark&&\checkmark&&\checkmark\\
Presto &2 years&1 month&100m$^2$&10m&\checkmark&\checkmark & \checkmark &\checkmark&\checkmark&&&&\\
ClimaX &1 week&6 hours&global&1.14°&&&&\checkmark&\checkmark &\checkmark&&&\\ 
Pritvi WxC &36 hours&3 hours&global&0.50°&&\checkmark&&\checkmark&\checkmark&\checkmark&\checkmark&&\\
Pangu-Weather &24 hours&1 hour&global&0.25°&&&&\checkmark&\checkmark&\checkmark&&&\\
GraphCast &18 hours&6 hours&global&0.25°&&&&\checkmark&\checkmark&\checkmark&&&\\
Aurora &18 hours&6 hours&global    &0.10°&&&&\checkmark&\checkmark&\checkmark&&&\\
FourCastNet &12 hours&6 hours&global&0.25°&&&&\checkmark&\checkmark&\checkmark&&&\\

\bottomrule
\end{tabular}
\caption{Comparison of FMs by property relevant for agricultural applications. Modalities included are: remote sensing (RS), including multispectral (MS), vegetation indices (VI) and Synthetic Aperture Radar (SAR); meteorological variables (Meteo), including temperature (T), precipitation (P) and additional modalities (+); other modalities including soil (S), farm management (M) and economic indicators (E). }
\label{tab:foundation_models}
\end{table*}

\section{Related Work}
\label{sec:relatedwork}

\subsection{Agricultural Tasks}
In this work, we concentrate on three important tasks for food security, shortly introduced below. 

\noindent\textbf{Crop Type Mapping.} While traditional crop type mapping methods rely on machine learning applied to designed temporal features based on vegetation indices \cite{QIU201735, Yang2019}, recent deep learning approaches leverage multi-temporal, multi-spectral data and foundation models pretrained on diverse modalities have improved accuracy, scalability, and generalization across regions while reducing the need for extensive labeled data \cite{10720850, Wang2025, Wu2025, Ruwurm2023, Ru_wurm_2020, tseng2023lightweight, tseng2025galileolearningglobal, chang2025generalizabilityfoundationmodelscrop}.

\noindent\textbf{Crop Yield Estimation.} Crop yield estimation has traditionally relied on process-based simulation and statistical models to link yield with environmental variables~\cite{Basso2019, vanDiepen1989, Jones2003, DSSAT_2019, Challinor2004, Shahhosseini2021, Jeong2022, Yang2023, Chang2023}. Recently, the availability of high-quality agricultural data has enabled machine learning and deep learning approaches that combine satellite imagery, weather, crop model outputs, and soil information to capture complex patterns and adapt to new environments~\cite{Maestrini2022, dilli_2022, Paudel2022, lin2023mmstvitclimatechangeawarecrop, Peng2024}.

\noindent\textbf{Crop Phenology Estimation.} Phenology is traditionally modeled using empirical biophysical model, such as temperature accumulation~\cite{Fishman1987} and chill requirements~\cite{Luedeling2010}, but often struggle to generalize across diverse environments due to variability in weather, soil, and landscape~\cite{Brown2008, Siebert2012}. Recent approaches integrate biophysical models with machine learning~\cite{vanbree2025hybridphenologymodelingpredicting} and combine multiple data sources, including remote sensing and meteorological data, to enhance accuracy and robustness~\cite{Lobert2023, Liao2023, Ma2023}.

\subsection{Foundational Models}
We categorize FMs based on their input features, namely \textit{Remote Sensing and Weather} and use the term \textit{FMs} to refer to models that have been trained on diverse data and that can be adapted to a range of downstream tasks \cite{bommasani2022opportunitiesrisksfoundationmodels, zhu2024foundationsearthclimatefoundation}. They are typically trained with self-supervised objective functions  on large volumes of data, often under a pretext task that enables the model to learn generalizable patterns.
\textbf{Remote sensing FMs} utilize satellite-derived data to extract features applicable to a variety of tasks such as land cover classification, crop type mapping, disaster assessment, and include models specialized in geo-location \cite{ayush2022geographyawareselfsupervisedlearning, klemmer2024satclipglobalgeneralpurposelocation}, temporal dynamics \cite{mañas2021seasonalcontrastunsupervisedpretraining}, multi-modal integration \cite{jakubik2025terramindlargescalegenerativemultimodality, sosa2025multimaemeetsearthobservation, nedungadi2024mmearthexploringmultimodalpretext, astruc2025anysatearthobservationmodel, tseng2023lightweight, tseng2025galileolearningglobal}, and multi-scale analysis \cite{reed2023scalemaescaleawaremaskedautoencoder}. 
\textbf{Weather-focused FMs} leverage large-scale meteorological datasets to provide efficient and scalable alternatives to traditional numerical weather prediction, enabling short- to medium-range atmospheric forecasting at global scales and varying resolutions \cite{nguyen2023climaxfoundationmodelweather, bodnar2024foundationmodelearth, nguyen2024scalingtransformerneuralnetworks, schmude2024prithviwxcfoundationmodel, lam2023graphcastlearningskillfulmediumrange}.
%
In Table~\ref{tab:foundation_models}, we qualitatively compare a range of general-purpose FMs on the properties of the ideal Crop FM. Our results are summarized, where we categorize models according to their primary category: MMEarth, AnySat~\cite{nedungadi2024mmearthexploringmultimodalpretext, astruc2025anysatearthobservationmodel} as \textit{Remote Sensing}; Galileo and Presto~\cite{tseng2025galileolearningglobal, tseng2023lightweight} as \textit{Remote Sensing and Weather}; and ClimaX, Prithvi WxC, Pangu-Weather, GraphCast, Aurora and FourCastNet ~\cite{nguyen2023climaxfoundationmodelweather, schmude2024prithviwxcfoundationmodel, bi2022panguweather3dhighresolutionmodel, lam2023graphcastlearningskillfulmediumrange, bodnar2024foundationmodelearth, pathak2022fourcastnetglobaldatadrivenhighresolution} as \textit{Weather} category.

\section{The Ideal Crop Foundation Model}
\label{sec:discussion}
Here, we first define the properties of an "ideal" FM for agriculture with current EO data. We consider properties of two categories: data modalities and spatiotemporal properties.

\textbf{Relevant modalities:}
For a FM to encode crop growth processes, it needs to include both the drivers and response of crop growth in its pretraining modalities. Mechanistic crop growth models such as APSIM, DSSAT and WOFOST~\cite{mccown1996apsim, jones2003dssat, van1989wofost} all utilize meteorology, soil and management as main drivers of crop response. Such data must be integrated into the CropFM model so it can, in principle, approximate similar processes to mechanistic crop growth models.

Gridded global datasets are available for the main meteorological drivers~\cite{Hersbach2020}, soil properties~\cite{poggio2021soilgrids} management~\cite{van2023worldcereal} and crop yield~\cite{Cao2025}. However, these gridded products are course-grained estimations and do not fully represent conditions at the farm scale. Closing this gap is essential, as drivers of crop growth can be scale-specific~\cite{rezaei2024impact}. In order to inform local growing conditions, remote sensing can be used. Multispectral and synthetic-aperture radar-based remote sensing contains information on crop type~\cite{Wu2025}, soil properties~\cite{Quast2023}, management including sowing~\cite{Lobert2023}, harvesting~\cite{Lobert2023}, irrigation~\cite{Boser2024}. Furthermore, vegetation indices are correlated with vegetation response, such as plant height~\cite{Tsao2023}, biomass~\cite{sialelli2025agbdglobalscalebiomassdataset}, growth stage~\cite{Liu2022} and plant health~\cite{johny2025bayesianhierarchicalframeworkfusion}. Finally, economical indicators, such as gross domestic product or nighttime light~\cite{elvidge2017viirs}, can indicate resources available for management actions and can explain the variation of crop growth across regions.

\textbf{Spatiotemporal properties:} 
We consider two spatiotemporal properties of FMs: the resolution and the extent. We define spatiotemporal resolution as the minimum interval in the input modalities across space and time, which corresponds to the shortest timestep and smallest pixel size in pretraining. Similarly, we define extent as the maximum interval across space and time within a single pretraining sample, which corresponds to maximum time series length and image area. We choose a spatial resolution of $\leq$10m and extent of $\geq$1 km$^2$ to cover a diverse spectrum of farm sizes, for which national averages range from approximately 0.03km$^2$ to 30 km$^2$~\cite{lowder2016number} globally. Crop growth has two natural time scales: a daily growth cycle, and the life-cycle of a crop, which ranges across crops from several months to several years (i.e. grasslands, perennial crops). As such, we aim for a temporal resolution of $\leq$1 day and a temporal extent of $\geq$1 year, with which we mirror the temporal resolution and extent of mechanistic crop growth models~\cite{mccown1996apsim, jones2003dssat, van1989wofost}. 

\section{Evaluating Current FMs}
In this preliminary evaluation, we demonstrate the potential of FMs in the agricultural tasks. 
Note, that this analysis presents work-in-progress results aimed to foster discussion at the workshop catalyzed by this paper:
~we evaluate Presto \cite{tseng2023lightweight} and Galileo \cite{tseng2025galileolearningglobal} on agricultural tasks, which are models particularly suitable for inputs with variable timesteps and point-based data, as they were trained on pixel-level timeseries. Additionally, their pretraining incorporates multiple modalities, including weather and soil data, which benefits agricultural applications.

\subsection{Agricultural Tasks}

\textbf{Crop Type Mapping.} Following the approach used in \cite{tseng2023lightweight}, we use the CropHarvest dataset \cite{tseng2021cropharvest}, focusing on maize classification in Kenya. We report the results directly as presented in \cite{tseng2023lightweight} and \cite{tseng2025galileolearningglobal}.

\noindent\textbf{Crop Yield Estimation.} For yield estimation, we utilize the CY-Bench dataset \cite{cybenchPaudel} and restrict our analysis to maize in Spain.  We compare our model’s performance to a random forest trained on all available features. The labels represent the end-of-season yield.

\noindent\textbf{Crop Phenology Estimation.} For phenology estimation, we obtain the tassel emergence date for maize from the Pan-European Phenology database \cite{Templ2018}, and download the input features (daily 2m temperature) from ERA5 reanalysis \cite{Hersbach2020}.  We restrict the region of evaluation to Germany.


\subsection{Experimental Setup}


For each downstream task, we evaluate three approaches: a Random Forest trained on (1) standard task-specific predictors (RF$^{TS}$), (2) FM-derived embeddings from Presto (Presto-RF) or Galileo (Galileo-RF), and (3) the raw input features provided to the FMs (RF$^{FM}$). This third comparison is necessary because the predictors typically used for each task may differ from the input formats required by FMs. 

In phenology estimation, the task-specific predictors are daily temperature, while the FMs generating the embeddings and RF$^{FM}$ use monthly-aggregated temperature as inputs. For yield estimation, the baseline Random Forest (RF$^{TS}$) employs a wider set of predictors as in \cite{cybenchPaudel}, whereas the embeddings used by Presto-RF, Galileo-RF, and the inputs to RF$^{FM}$ are based solely on temperature and precipitation.

Yield forecasting applies leave-one-year-out cross-validation over 18 years of data. Phenology estimation uses 33 years of data, with the last two years for testing. Crop type mapping follows the data splits of \cite{tseng2023lightweight}.

\subsection{Results}

Table \ref{tab:presto_combined} shows that both Presto-RF and Galileo-RF outperform the baseline RF$^{TS}$ in crop type mapping, likely due to the multi-modal pretraining of the FM embeddings and their ability to extract robust, generalizable features. For yield estimation, however, RF$^{TS}$ and RF$^{FM}$ achieve lower RMSE values—even when restricted to temperature and precipitation—indicating that FM-derived embeddings do not provide additional useful information for this task when only limited predictors are available. In phenology estimation, Presto-RF, Galileo-RF, and RF$^{FM}$ perform comparably when evaluated on monthly aggregated temperature data, highlighting the effectiveness of diverse pretraining in this context. 

\begin{table}[t]
\centering
\footnotesize
\begin{tabular}{lccc}
\toprule
\textbf{Model}& \small{\textbf{Crop Type}} & \small{\textbf{Yield}} & \small{\textbf{Phenology}} \\
 & \textbf{F1 Score} $\uparrow$ & \textbf{NRMSE} $\downarrow$ & \textbf{RMSE (days)} $\downarrow$ \\

\midrule
RF$^{\text{TS}}$ & 0.559 & \textbf{14.28} & 10.19 \\
RF$^{\text{FM}}$  & ---   & 26.59 & \textbf{9.19} \\
Presto-RF           & 0.840 & 31.10 & 9.82 \\
Galileo-RF          & \textbf{0.845} & 29.32 & 9.41 \\
\bottomrule
\end{tabular}
\caption{Performance comparison in three agricultural tasks: crop type mapping (Kenya), yield estimation (Spain), and phenology estimation (Germany). RF$^{\text{TS}}$, RF$^{\text{FM}}$, Presto-RF, and Galileo-RF are Random Forest models trained on task-specific inputs, FM inputs, and FM-derived embeddings, respectively.
}
\label{tab:presto_combined}
\end{table}



\section{Discussion}

A key technical challenge is that most existing model architectures lack the granularity and adaptability necessary for point-based agricultural tasks. For remote sensing, transformer-based models typically pretrain on fixed-size patches; while some, like AnySat with FlexiViT \cite{beyer2023flexivitmodelpatchsizes}, support dynamic patching, most cannot handle variable-sized regions without retraining. Feeding single patches to standard ViTs is possible but often ineffective, as these models rely on spatial context learned during pretraining. For weather models, grid-based architectures \cite{bodnar2024foundationmodelearth, nguyen2023climaxfoundationmodelweather} operate at coarse resolutions (0.1–5.626 degrees), so each grid cell covers many farms, making direct application to fine-scale tasks impractical without down-scaling. Models like Presto and Galileo address this partially by using pixel-level timeseries, but their utility depends on how well the pixel’s physical meaning matches the task requirements. 

We identified Presto and Galileo as FMs with the highest suitability for agricultural applications. However,  our case study suggest that they cannot be directly applied across agricultural tasks. Based on Table \ref{tab:foundation_models}, two main limitations emerge: insufficient temporal resolution, and the absence of key agriculturally relevant modalities. 

Increasing the temporal resolution to daily would align the FMs with the timescale of current process-based crop models ~\cite{mccown1996apsim,jones2003dssat,van1989wofost} and broaden their applicability, as crop response to extreme meteorological events and  management actions are captures a daily timescale. While our results for phenology estimation (Table \ref{tab:presto_combined}) suggest that increasing the temporal resolution of input data from monthly to daily with a limited number of training samples can be detrimental to baseline methods, FMs---pretrained on several orders of magnitude more data---should be less prone to overfitting at higher resolutions. 

Expanding the range of pretraining modalities is also crucial. Our yield estimation results (Table \ref{tab:presto_combined}) reveal that restricting inputs to those used by Presto significantly reduces performance, indicating these modalities are insufficient for accurate yield prediction. 
Fortunately, current FMs are designed for multimodal integration, hence adding more vegetation indices, meteorological variables, soil properties and socio-economical indicators is feasible. Although high-quality RS data could  compensate for missing modalities, it remains unclear if this potential is fully utilized during pretraining.

Additionally, integrating large language models (LLMs) with FMs can enrich agricultural applications by incorporating textual data from manuals and reports, capturing management practices often absent from sensor data \cite{Tzachor2023, Peng2023, Li2024}. LLMs also offer practical tools for extension services, translating complex scientific knowledge into personalized, data-driven advice for farmers at scale \cite{Silva2023_a, Tzachor2023, DeClercq2024}. 

\section{Conclusion}

In this study, we evaluated two FMs on three large-scale, food security–relevant agricultural tasks for maize: crop type mapping in Africa, yield estimation, and phenology estimation in Europe. Random Forests with FM-derived embeddings outperformed task-specific models in crop type mapping, were comparable in phenology, but lagged in yield estimation likely due to limited input variables. These findings highlight both the promise and current challenges of adapting FMs to agricultural tasks, particularly when pretraining inputs differ from task requirements. Our results and recent advances suggest that FMs could help address domain shift and deliver actionable, data-driven insights to support agricultural decision-making. As with any deep learning system, a universally applicable CropFM will need to incorporate both design knowledge in terms of model architecture and learning objective and and agricultural domain knowledge through an informed choice of input modalities and task design. Architecture-wise, further development in the incorporation of heterogeneous temporal and spatial resolutions will be necessary, while integrating fine-grained environmental data is key for CropFM models to provide the data-foundation to approximate the complex internal biological processes in vegetation.

In terms of future potential, FMs hold great promise for agriculture by addressing domain shift through pretraining on large, diverse datasets encompassing remote sensing, climate, soil, agricultural management and socio-economic data. This broad training enables FMs to capture spatial and temporal variability in crop growth, improving transferability and robustness across regions and time periods \cite{Ma2024, He2023, chen2023foundationmodelsweatherclimate, lu2025visionfoundationmodelsremote, Li2024}.

\clearpage 
{
    \section*{Acknowledgments}
    This work was partially supported by the Digital Europe Programme under Grant agreement \href{http://www.agrifoodtef.eu}{AgrifoodTEF} - Test and Experiment Facilities for the Agri-Food Domain (Grant \#\href{https://ec.europa.eu/info/funding-tenders/opportunities/portal/screen/opportunities/projects-details/43152860/101100622}{101100622})

    \small
    \bibliographystyle{ieeenat_fullname}
    \bibliography{main}
}

\end{document}